\documentclass[sigconf]{acmart}
\usepackage{enumerate}
\usepackage{graphicx}
\usepackage{subcaption}

\usepackage{amsmath, amssymb}
\usepackage{multirow}
\usepackage{acronym}
\usepackage{balance}

\AtBeginDocument{%
  }

\copyrightyear{2025}
\acmYear{2025}
\setcopyright{acmlicensed}
\acmConference[SIGIR '25] {Proceedings of the 48th International ACM SIGIR Conference on Research and Development in Information Retrieval}{ July 13--18, 2025}{Padua, Italy.}
\acmBooktitle{Proceedings of the 48th International ACM SIGIR Conference on Research and Development in Information Retrieval (SIGIR '25), July 13--18, 2025, Padua, Italy}

\acrodef{CPS}[CPS]{Conversational Product Search}

\begin{document}

\title{PSCon: Product Search Through Conversations}


\author{Jie Zou}
\affiliation{%
 \institution{University of Electronic Science and Technology of China}
   \city{Chengdu}
  \country{China}
 }
\email{jie.zou@uestc.edu.cn}

\author{Mohammad Aliannejadi}
\affiliation{%
  \institution{University of Amsterdam}
   \city{Amsterdam}
  \country{The Netherlands}
}
\email{m.aliannejadi@uva.nl}

\author{Evangelos Kanoulas}
\affiliation{%
  \institution{University of Amsterdam}
   \city{Amsterdam}
  \country{The Netherlands}
}
\email{ekanoulas@gmail.com}

\author{Shuxi Han}
\affiliation{%
  \institution{University of Electronic Science and Technology of China}
   \city{Chengdu}
  \country{China}
  }
\email{501560987@qq.com}

\author{Heli Ma}
\affiliation{%
  \institution{University of Electronic Science and Technology of China}
   \city{Chengdu}
  \country{China}
  }
\email{1357137816@qq.com}

\author{Zheng Wang}
\affiliation{%
  \institution{Tongji University}
   \city{Shanghai}
  \country{China}
  }
\email{zh_wang@hotmail.com}

\author{Yang Yang}
\authornote{Corresponding author.}
\affiliation{%
  \institution{University of Electronic Science and Technology of China}
   \city{Chengdu}
  \country{China}
  }
\email{yang.yang@uestc.edu.cn}

\author{Heng Tao Shen}
\affiliation{%
  \institution{University of Electronic Science and Technology of China}
   \city{Chengdu}
  \country{China}
  }
\email{shenhengtao@hotmail.com}

\renewcommand{\shortauthors}{Jie Zou et al.}

\begin{abstract}
\ac{CPS} systems interact with users via natural language to offer personalized and context-aware product lists. However, most existing research on \ac{CPS} is limited to simulated conversations, due to the lack of a real \ac{CPS} dataset driven by human-like language. Moreover, existing conversational datasets for e-commerce are constructed for a particular market or a particular language and thus can not support cross-market and multi-lingual usage. In this paper, we propose a CPS data collection protocol and create a new \ac{CPS} dataset, called PSCon, which assists product search through conversations with human-like language. The dataset is collected by a coached human-human data collection protocol and is available for dual markets and two languages. By formulating the task of \ac{CPS}, the dataset allows for comprehensive and in-depth research on six subtasks: user intent detection, keyword extraction, system action prediction, question selection, item ranking, and response generation. Moreover, we present a concise analysis of the dataset and propose a benchmark model on the proposed \ac{CPS} dataset. Our proposed dataset and model will be helpful for facilitating future research on \ac{CPS}.

\end{abstract}



\maketitle

\section{Introduction}

In traditional product search systems, users formulate queries and then browse the products to locate their target items (i.e., products). This is, however, inefficient and may be mismatched by product search systems because of the semantic gap between queries and products~\cite{van2016learning,zou2019learning}. Recently, researchers have augmented search functionality by allowing search systems to interact with users by natural language and collect explicit feedback from users as a step towards better understanding users' information needs~\cite{zou2019learning,bi2019conversational,zhang2018towards}.

As for the existing research on Conversational Product Search (CPS), \citet{zhang2018towards} presented the first CPS model by simulating conversations over item ``aspects'' extracted from user reviews. Based on that, \citet{bi2019conversational} proposed a \ac{CPS} model based on the negative feedback and simulated user feedback on the item ``aspects.'' Similarly, \citet{zou2022learning} simulated conversations based on item ``aspects'' and proposed a \ac{CPS} model via representation learning. On the other hand,~\citet{zou2019learning} proposed a question-based Bayesian product search model by simulating conversations based on extracted informative terms instead of item ``aspects.'' 
The existing work on \ac{CPS} mainly simulates user conversations while interacting with the product search system. That is, there is a lack of a real \ac{CPS} dataset driven by human-like language, and thus a public dataset for product search with human-like conversations is of great importance and significance. 

Moreover, most conversational domain-specific datasets are created for a \textit{particular market} or a \textit{particular language}. However, (i) e-commerce companies often operate across markets. For example, the Amazon platform 
has 20 markets around the world\footnote{https://sell.amazon.com/global-selling.html} for global selling. Typically, some e-commerce companies operate in multiple countries and they can benefit from the experience and data gathered across several markets. Therefore, it is significant to construct a cross-market dataset for product search~\cite{bonab2021cross}; (ii) different users from different countries use different languages and each language has its own grammar and syntactic rules. It is important to create a non-single language \ac{CPS} dataset to facilitate the development of \ac{CPS} models that can generalize well to different languages. Yet resources of conversational datasets in languages other than English are lacking. 

To this end, in this paper, we create PSCon, a new dataset for product search with human-like conversations that supports dual markets and two languages. Specifically, we build the dataset by formulating the task of \ac{CPS}. We define a pipeline for \ac{CPS}, which includes the following six sub-tasks: (T1) user intent detection, (T2) keyword extraction, (T3) system action prediction, (T4) question selection, (T5) item ranking, and (T6) response generation. First, user intent detection aims to identify the user's general intent. Second, for each utterance from the user, the system needs to understand its fine-grained information by extracting keywords. Third, the system needs to decide which action to take at each turn, e.g., whether the system should ask a clarifying question or return products to users. Fourth, when the system decides to ask a clarifying question, question selection focuses on selecting high-quality questions to ask the user for preference elicitation. Fifth, when the system decides to return products to users, the system selects products based on conversation history and returns the high-quality products to users. Sixth, response generation incorporates useful information and generates natural language responses to communicate with the user. To collect the \ac{CPS} dataset, we use a coached human-human data collection protocol to ensure human-like dialogues and avoid algorithm bias~\cite{radlinski2019coached}, where some participants mimic system roles, i.e., digital shopping assistants, while others act as user roles, i.e., customers, similar to the setting of existing conversational datasets ~\cite{bernard2023mg}. The user role expresses their demands for products and the system role assists them in finding their target products through conversations. 

In summary, the main contributions of this work are fourfold: 
\begin{itemize}
    \item We propose a protocol to collect the \ac{CPS} dataset and design a tool to perform data collection following our protocol. 
    \item We formulate a pipeline for \ac{CPS}, which  consists of six sub-tasks. 
    \item We build a CPS dataset through human-human conversations, named PSCon, that supports dual markets and two languages. We also build a knowledge graph to support the development of knowledge graph-based \ac{CPS} models.
    \item We present a concise analysis of the dataset and propose a benchmark model for \ac{CPS}. 
\end{itemize}
 To the best of our knowledge, this is the first effort for the \ac{CPS} dataset with human-like language that is available for dual markets and two languages. The dataset and code are available on https://github.com/JieZouIR/PSCon.

\section{Related Work}
\subsection{Conversational Product Search}
Recently, there has been a growing interest regarding \ac{CPS} in the information retrieval community \cite{javadi2023opinionconv, montazeralghaem2022learning, papenmeier2023ah, ye2024productagent, li2025wizard}. In the early stage,~\citet{zhang2018towards} proposed a first \ac{CPS} model by introducing a unified framework for conversational search and recommendation. 
This model engages users by posing questions about various aspects derived from user reviews and collects their feedback on these aspect values. Similarly,~\citet{bi2019conversational} simulated conversations and solicited explicit user responses to aspect-value pairs mined from product reviews, with a particular emphasis on the importance of negative feedback in the context of \ac{CPS}. In contrast to this approach,~\citet{zou2019learning} focused on querying users about informative terms, typically entities, extracted from item-related descriptions and reviews. They introduced a sequential Bayesian method that employs cross-user duet training for enhancing \ac{CPS}. 
Based on that, ~\citet{zou2020empirical} and ~\citet{ma2024ask} conducted empirical studies to measure user interactions in question-based product search systems and \ac{CPS}, shedding light on the effectiveness of existing question-based product search approaches. 
More recently,~\citet{zou2022learning} asked clarifying questions \cite{vedula2024question} on item aspects and proposed a \ac{CPS} model via representation learning. 

The aforementioned studies on \ac{CPS} all simulate user conversations when interacting with the product search system. That is, their conversations are based on template-based questions and simulated user answers. Although they demonstrate CPS is a promising research direction,
their deployment of simulated conversations is suboptimal as this is not a human-like setting~\cite{zou2022learning}. This is mainly due to the lack of real conversations for product search. In this work, we fill this gap by proposing a dataset for \ac{CPS} with real human-like conversations. 

\subsection{Conversational Datasets}
There are several conversational datasets available for different tasks, such as conversational recommendation~\cite{moon2020situated,saha2018towards,wu2022state}, and conversational search~\cite{ren2021wizard}. The majority of conversational recommendation datasets focus on the movie domain, which involves a collection of annotated dialogs where a seeker requests movie suggestions from the recommender~\cite{zhou2020towards,li2018towards,liu2021durecdial}. In contrast, conversational search datasets~\cite{ren2021wizard,qu2018analyzing,nguyen2016ms,dalton2020cast,mcdumisc,aliannejadi2019asking,zamani2020mimics,aliannejadi2020convai3,chu2022convsearch,trippas2020towards,liao2021mmconv} focus on conversations to assist information seeking in the search scenarios. For instance, \citet{dalton2020cast} and \citet{thomas2017misc} released the CAsT and MISC datasets, which are created by volunteers or experts with 80 and 110 conversations, respectively. 
~\citet{ren2021wizard} proposed a conversational information-seeking dataset with the Wizard-of-Oz (WoZ) setup. In this paper, we use a human-human data collection protocol based on participants recruited from universities, which is more reliable \cite{bernard2023mg}. 
Compared with conversational search for locating relevant documents, product search focuses on locating the potential products for purchase~\cite{wu2018turning,ai2017learning}, which is a subset of relevant products and thus more challenging. Moreover, user queries in product search are often short and thus preference elicitation plays an important role in \ac{CPS} due to its goal of locating the purchased products. 

As for product search and e-commerce, there are only a few conversational datasets available. \citet{fu2020cookie} built a conversational dataset for e-commerce constructed from Amazon reviews. However, they use simulated users and conversations. \citet{jia2022convrec} introduced a conversational dataset from the E-commerce domain with user profiles, but they are in Chinese only and involve a single market only. \citet{bernard2023mg} collected a dataset of multi-goal conversations for e-commerce, which is the closest dataset to ours. However, they are in a single language, i.e., English, and involve a single market only. Also, they only contain less than 20 conversations per category. Different from the above publications, we build a \ac{CPS} dataset through real human-human conversations, which is on a suitable scale. Moreover, it can support dual markets and two languages, along with a knowledge graph.

\subsection{Conversational Recommender System for e-Commerce} Conversational recommender systems for e-commerce are also associated with product search. For instance,~\citet{zhang2018towards} propose a unified paradigm for product search and recommendation. Conversational recommender systems utilize human-like natural language to deliver personalized and engaging recommendations through conversational interfaces like chatbots and intelligence assistants \citep{gao2021advances,jannach2021survey}. In general, the existing research on conversational recommender systems for e-commerce is usually for anchor-based conversational recommender systems, 
i.e., they simulate conversations based on the predefined anchors to characterize items, including intent slots (e.g., item aspects and facets)~\citep{zhang2018towards}, entities~\citep{zou2020towardsb}, and attributes~\citep{zhang2022multiple}. In contrast to the existing work for conversational recommendation that mainly focuses on preference elicitation and top-k recommendation~\cite{radlinski2019coached,sun2018conversational}, our \ac{CPS} focuses on item ranking and incorporates user intent detection, keyword extraction, system action prediction, and response generation. That is, they only cover part of the aspects of the \ac{CPS} pipeline. 

\begin{figure}[t]
\centering
\includegraphics[width=\columnwidth]{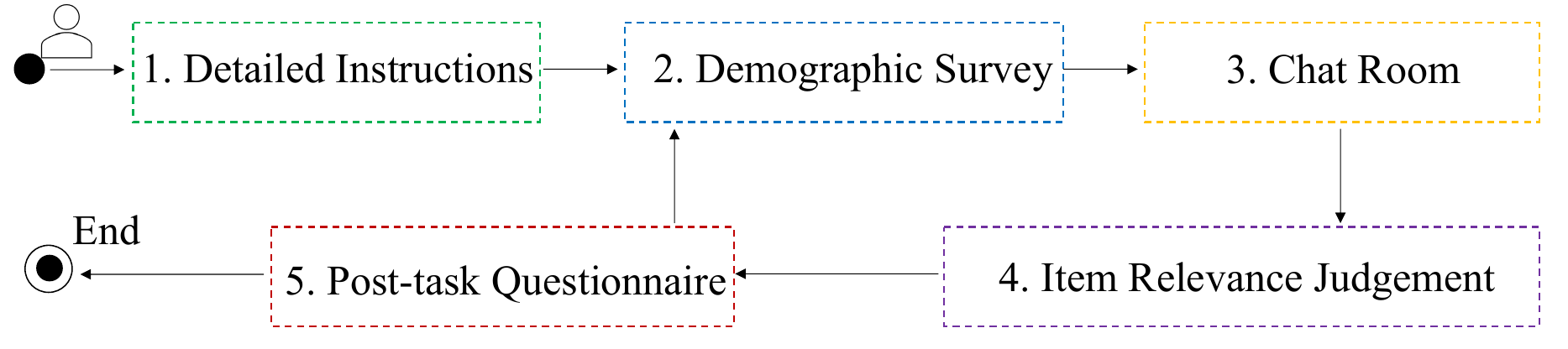}
\caption{Study protocol overview.}
\label{fig:pipline}
\end{figure}

\section{Data Collection \& Analysis}
\label{sec:dat}
\subsection{Protocol}
We have developed an online Web system to collect the data. Following Figure~\ref{fig:pipline}, the human-human data collection protocol used in this study is as follows:
\begin{enumerate}[(1)]
    \item Participants are presented with detailed instructions regarding this study. They are also trained to be familiar with the system through detailed videos. 
    \item Participants are asked to complete a demographic survey, e.g., their gender, age, and career field.
    \item The study uses a human-human data collection protocol, where some participants mimic system roles to help users locate their target products, while others play user roles to seek products. They chat in the chat room. 
    \item The conversation starts from the user role. From the beginning, we ask the user role to imagine a target product that she/he wants to buy in mind. Then, the user role initiates the conversation by revealing the product needs and chatting with the system to get the target product. 
    \item The user role is asked to pick a search intent (e.g., uttering a greeting or sending a request (Section~\ref{sec:intent})) and then start the conversation by writing a message to send to the system. 
    \item Whenever the system role receives the message from the user role, the system role needs to extract keywords (Section~\ref{sec:slotvalue}) from the conversation history that are used to understand the natural language conversation. 
    After that, the system role has to select an action label that reflects the action to take (Section~\ref{sec:action}), e.g., asking a clarifying question (Section~\ref{sec:Qselect}) or recommending products (Section~\ref{sec:itemselect}). 
    At last, the system role needs to generate a natural language response to send to the user based on the selected action accordingly (Section~\ref{sec:response}).
    \item For each round of conversation, the system roles can input a query (i.e., keywords) to search for related results for products. They can also search for products by using the search filters provided in the system when necessary. 
    \item When the system roles think they need more information about the user's need, they can scan corresponding attributes of products provided by the Amazon product search API, and ask a clarifying question about any attributes (or others) to the user. 
    \item When the system roles think that there are products meeting the user's need, they can select corresponding products from the search result panel (based on the Amazon product search API) to recommend to the user.
     \item The conversation repeats until the user ends the chat. For each turn, both the user and system role can send multiple messages at once. For messages, they can send text, product images, or URLs. 
     \item After the conversation is done, we ask the user role to mention if the system managed to find their target products or not. The user is also asked to evaluate the relevance of the products recommended by the system in the conversations. 
     \item At the end, participants are asked to complete a post-task questionnaire, on their experience and satisfaction.
\end{enumerate}
We value participants' privacy. All participants are informed that their data is securely encrypted and any identifiable information is removed. We did not collect any data to breach their privacy. We also ask the participants to report any adult content and/or offensive content in the conversations that might raise ethical issues. 

\begin{figure*}[tb]
\centering
\begin{subfigure}{0.228\textwidth}
    \includegraphics[width=\textwidth]{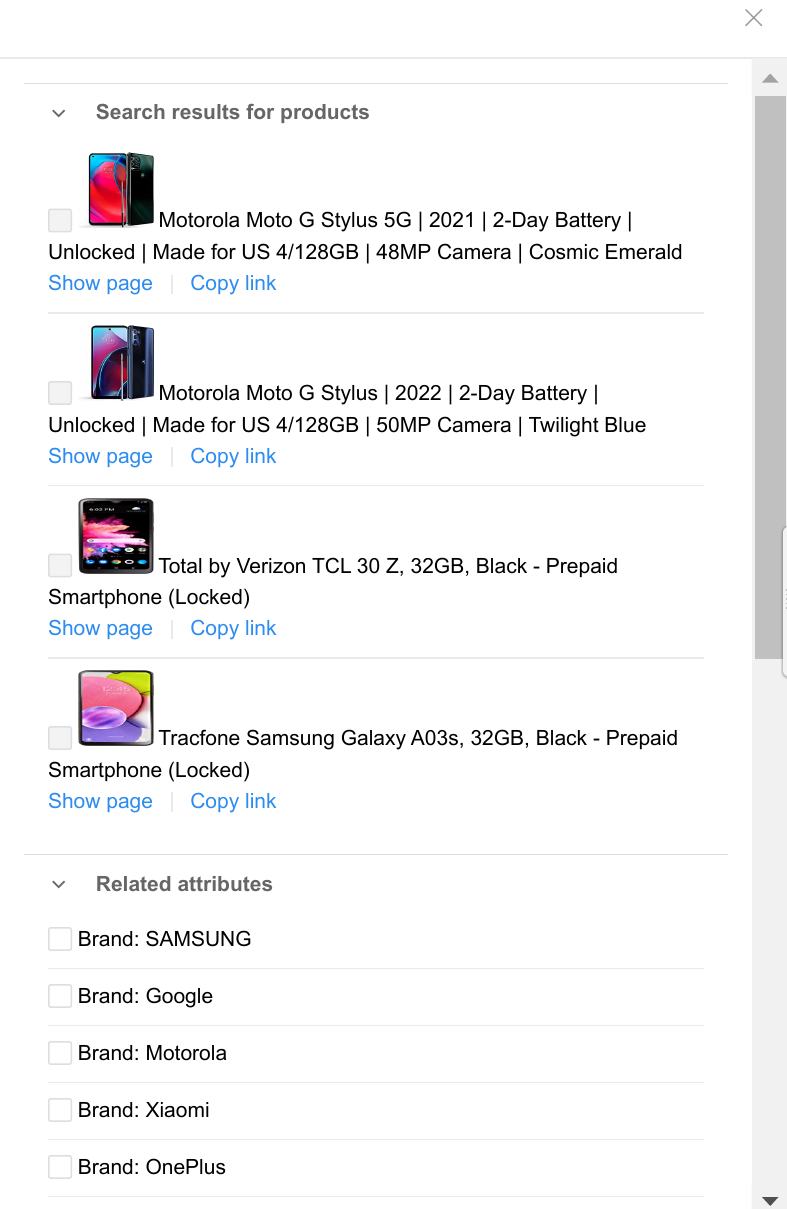}
    \caption{Search interface.}
\end{subfigure}
\begin{subfigure}{0.62\textwidth}
    \includegraphics[width=\textwidth]{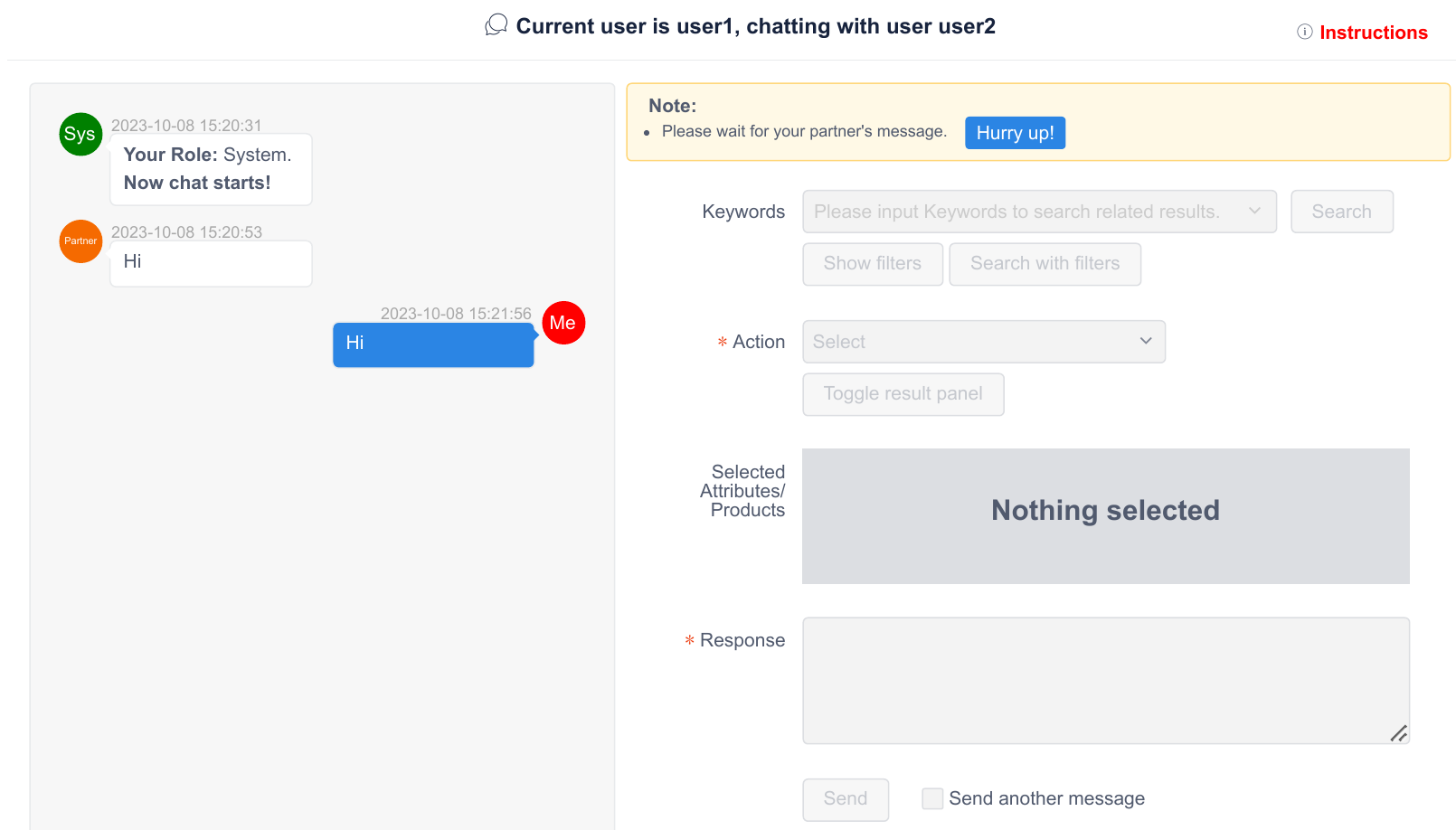}
    \caption{Chat interface.}
\end{subfigure}
\caption{The interface of the chat room (the system role).}
\label{fig:sysinterface}
\end{figure*}

\subsection{Data Collection System}
The online Web system for data collection mainly includes administration pages and the chat room. 

\subsubsection{Administration Pages}
When entering the system, the new participant needs to register first. When registration, the new participant is assigned a role (i.e., the user role or the system role) randomly, following the random assignment setup of related literature~\cite{HarveyP17,kelly2009methods,zou2022Users}. After logging in with the assigned role, participants are directed to the instruction page. The instruction page shows detailed instructions regarding the study and what is expected concerning conversational goals. The instructions are designed for each role. That is, the participants of each role only see their own instructions for the customers or shopping assistants. 
Following the instructions, some demographic questionnaires are shown to the participants, such as their gender, age, and career field. After that, the participants are redirected to the chat room (see Section \ref{para:chatroom} for details). After the conversation in the chat room is done, the participants are redirected to a post-task page. The post-task page shows some post-task questionnaires regarding the experience and satisfaction of participants. In addition, there is a relevance judgment page before the post-task questionnaires. On the relevance judgment page, the user role needs to evaluate the relevance of the products (using like/dislike labels to indicate their assessment of the recommended items) recommended by the system in the conversations.

\subsubsection{Chat Room}
\label{para:chatroom}
The interface for the chat room differs depending on the role of the participant. The interface for the system role is shown in Figure~\ref{fig:sysinterface}, while the interface for the user role is omitted due to its simplicity and limited space. 
Both the interface for the user role and the system role are divided into two vertical blocks. The left block displays the ongoing conversation while the right block shows the control panel. They both contain an instruction in the top right corner. 
The interface for the control panel differs between the user role and the system role. For the user roles, the control panel allows them to select an intent and send messages. They can also finish the conversation anytime by clicking the ``finish the conversation" button at the bottom. For the system roles, the control panel also allows them to select a system action and send messages. In addition, the system interface includes a search engine. The search engine produces results of products and attributes by keywords. The system role is allowed to submit keywords to search results from the search engine. The search engine we used in this paper is based on the Amazon product search API.\footnote{We use Amazon China (https://www.amazon.cn/) when collecting the Chinese dataset, while using Amazon USA (https://www.amazon.com/) when collecting the English dataset.}
During the conversation, the system allows them to select products from the entire Amazon product pool to send to the user. The selected product list is comprised of curated products with descriptions and pictures. Both the user and the system can urge their chat partner to speed up, which is shown on the upper right of the interface. 

\subsection{Quality Control}
To ensure data quality, we perform quality control measures to avoid including low-quality conversations. Specifically, we deploy five quality checks: (1) we perform an on-site training on the study goal and how to use the system for all participants before the participants start the study; (2) we provide detailed videos demonstrating the process and detailed documentation on the study, e.g., the instruction of using the interfaces, the steps in creating the data, the explanation of the labels, and the positive and negative examples; (3) we evaluate the time participants spent reading the textual descriptions and ask participants questions about the study descriptions to ensure that they have read and understood the instructions; (4) we remove the data with incomplete conversations (i.e., not ended normally or conversations without any products recommended); (5) The collected data are further checked manually to ensure that removed participants and conversations are not filtered out wrongly. We also manually go over the data and correct issues such as typos, grammatical errors, and mislabels.\footnote {We conducted a detailed analysis for user satisfaction on the dataset. We found that, in 97.2\% of cases, users successfully found their target products, and users provided an average satisfaction score of 4.91 (scale of 1 to 5) for the system's recommendations. So this indeed suggests that users were satisfied with the system's recommendation and confirms the applicability of our dataset.}

\subsection{Participants} 
\label{sec:participants}
Participants in this study were 465 volunteers recruited through email invitations (students and staff of two universities, one in Europe and one in Asia). Of those participants, 63 were in the pilot study and 402 were in the actual study. The pilot study is deployed to make sure the data collection runs smoothly as expected before the actual data collection. 
Participants were paid around 2.5 pounds on average for each conversation. 
In the actual study, 196 participants took part in the Chinese dataset collection and 206 participants took part in the English dataset collection. 

\begin{table}[t]
\caption{Statistics of the collected datasets.}
\label{tab:data}
\centering
  \small
\begin{tabular}{ll|ll}
\toprule
\multicolumn{2}{c|}{\textbf{Chinese Dataset}} & \multicolumn{2}{c}{\textbf{English Dataset}}\\
\midrule 
\# conversations & 904 & \# conversations & 826\\
\# utterances & 5,254 & \# utterances & 5,633\\
\# turns & 3,044&\# turns & 3,187\\
\# items & 2,548&\# items & 2,664\\
avg. \# utterances & 5.81&avg. \# utterances & 6.82\\
avg. \# turns & 3.37&avg. \# turns & 3.86\\
avg. \# items & 2.82&avg. \# items & 3.23\\
\bottomrule
\end{tabular}
\end{table}

\subsection{Data Statistics}
The datasets contain an English and a Chinese dataset, which include a total of 1,730 conversations and 10,887 utterances recommending 5,212 products. The data statistics for the datasets are reported in Table ~\ref{tab:data}. The Chinese dataset includes a total of 904 conversations and 5,254 utterances recommending 2,548 products. Each conversation contains 5 to 16 utterances, 3 to 7 turns, and 1 to 30 products. 98.01\% of the conversations have 3-4 turns. 91.82\% of the conversations recommend 1-4 products. The English dataset includes a total of 826 conversations and 5,633 utterances recommending 2,664 products. Each conversation contains 2 to 17 utterances, 1 to 8 turns, and 1 to 14 products. 96.58\% of the conversations have 2-6 turns. 85.98\% of the conversations recommend 1-6 products. This is in line with that users expect the system to recommend high-quality products with fewer rounds in real applications, and thus they prefer to converse with short interactions by following such a real-world setting.\footnote{To try our best to mimic a real-world scenario, we ask the user to imagine an initial target product (either specific, ambiguous, or evolving product needs) to buy in mind for product needs and do not set restrictions for long conversations and open dialogues. Therefore, users can discuss products as they like, similar to a normal user in a real-world product search system.} For the Chinese dataset, the average number of utterances per conversation is 5.81, the average number of turns per conversation is 3.37, and the average number of items per conversation is 2.82. On average, utterances have 8.17 $\pm$ 4.43 words. For the English dataset, the average number of utterances per conversation is 6.82, the average number of turns per conversation is 3.86, and the average number of items per conversation is 3.23. On average, utterances have 6.57 $\pm$ 5.12 words. 

\begin{figure}[t]
\centering
\includegraphics[width=0.85\columnwidth]{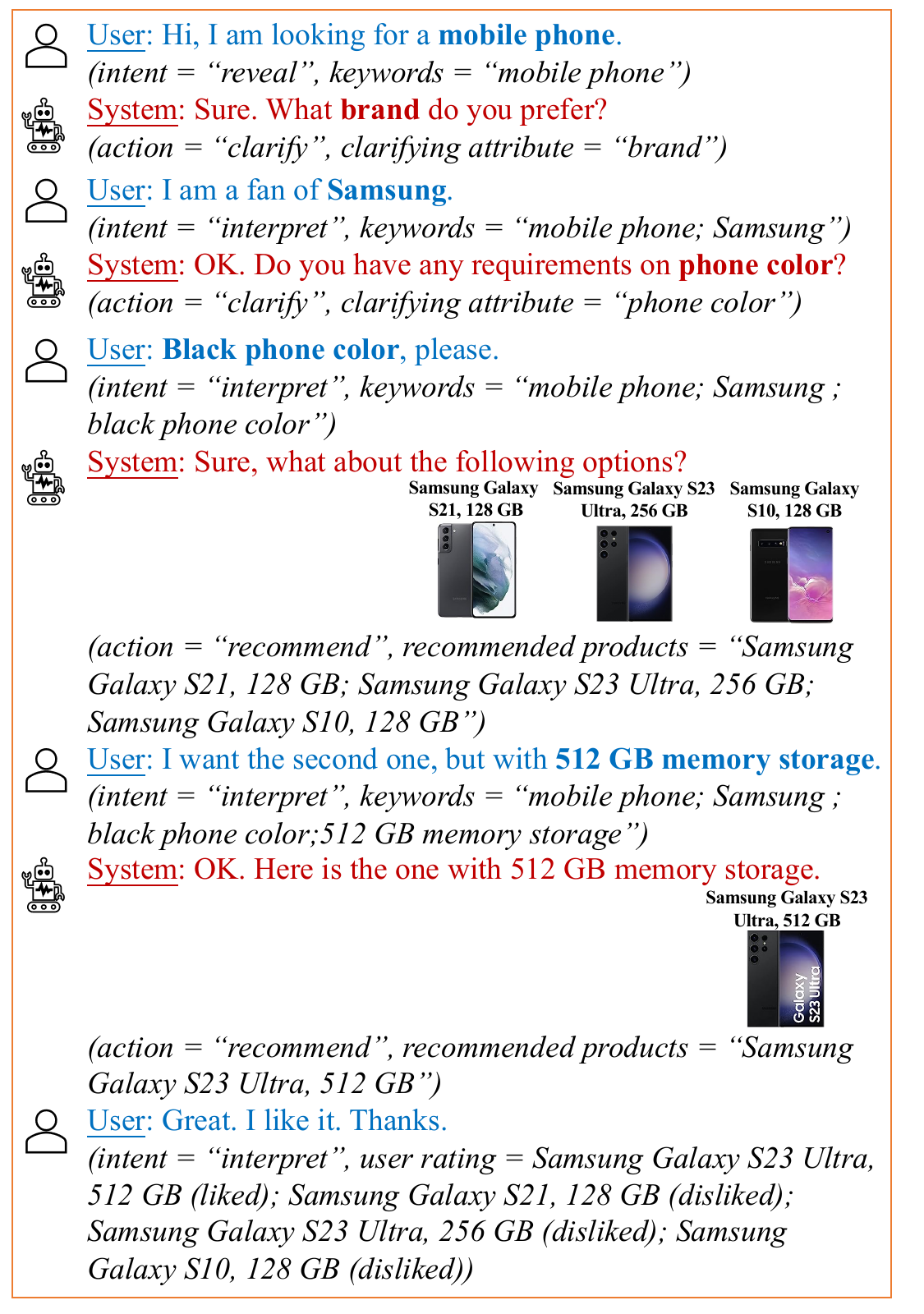}
\caption{A conversation example of our collected dataset.}
\label{fig:example}
\end{figure}

Besides the conversation, we also collect the information related to the conversation. We collect the conversation annotations, including the user intent, keywords, system action, attributes for clarifying questions, and selected products recommended in the conversation to support our subtasks.\footnote{The annotations (ground truth in the dataset) are provided by the recruited participants who act as the system roles or user roles. They provide annotations when they chat with each other, for each round of conversation. } The example of a collected conversation is shown in Figure~\ref{fig:example}. Also, the user preferences for products (like/dislike) recommended by the system in the conversation is collected. For each search behavior, the search results for products (top 50 products) and related attributes are also collected. In addition, we also collect information on each product including the URL, product image, product title, product description, product reviews, and product metadata such as category, brand, features, product ratings, and related products (also bought, also viewed, bought together, compared, and sponsored products). Given that many existing studies on product search incorporate external knowledge graphs to capture user preference, we construct a knowledge graph by using the product metadata (i.e., users, products, like/dislike relationships between users and products; products, product attributes, and relationships between products and corresponding attributes) to support the development of \ac{CPS} models based on knowledge graphs.

\subsection{Data Analysis}

\begin{figure}[t]
\centering
\begin{subfigure}{0.23\textwidth}
    \includegraphics[width=\textwidth]{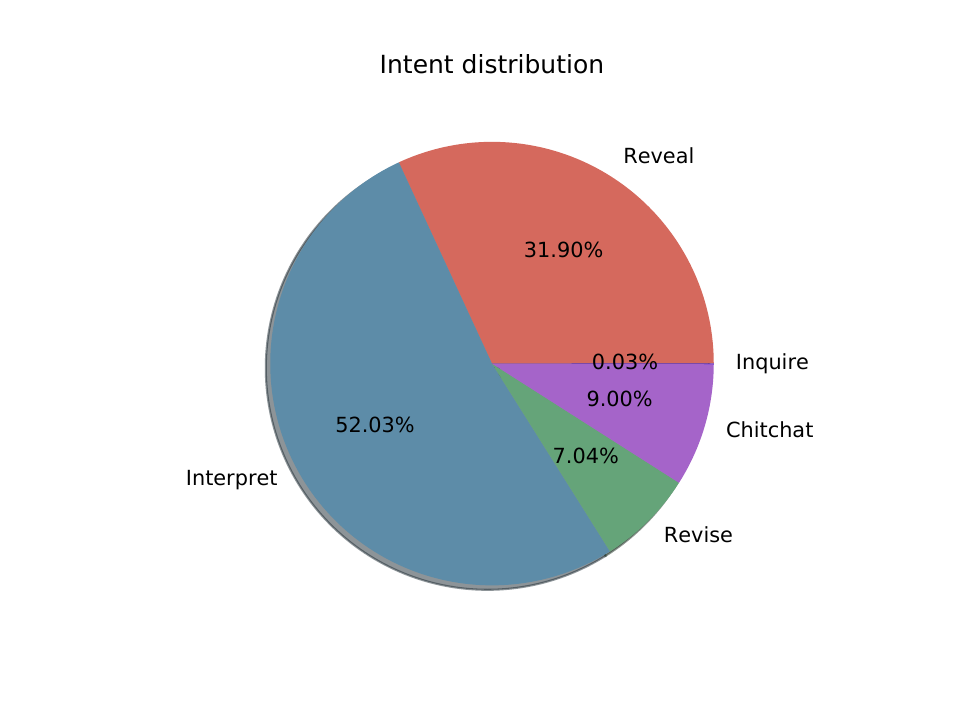}
    \caption{Intent distribution}
\end{subfigure}
\begin{subfigure}{0.24\textwidth}
    \includegraphics[width=\textwidth]{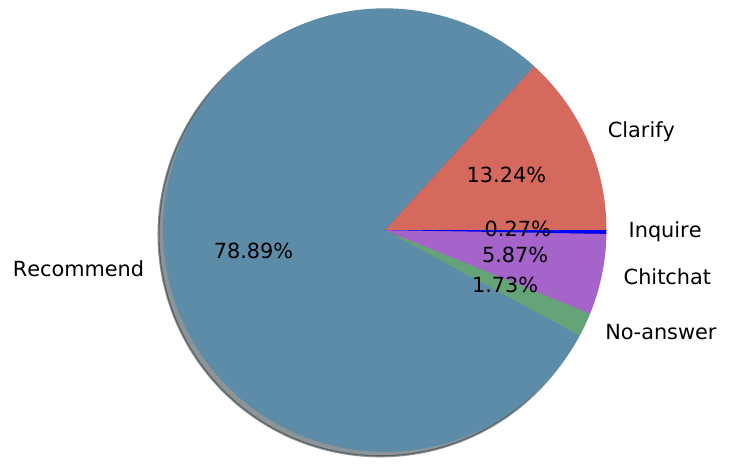}
    \caption{Action distribution.}
\end{subfigure}
\caption{Intent/action distribution in the Chinese dataset.}
\label{fig:pie}
\end{figure}

We analyze the PSCon dataset with a focus on the distribution of user intents and system actions, as shown in Figure~\ref{fig:pie}.\footnote{The distribution of intent/action is calculated as the percentage of each corresponding intent/action, i.e., the number of utterances with the corresponding intent/action labels divided by the total number of utterances with any intent/action labels.} We observe, in the Chinese dataset, the top intents selected by users are ``Interpret'' and ``Reveal,'' while the top system actions are ``Recommend'' and ``Clarify.'' This indicates that systems ask clarifying questions or recommend products in most cases, whereas users reveal their product needs and provide feedback to the system in most cases when interacting with the \ac{CPS} system. We omit the figures of the English dataset, due to a similar trend and limited space.

\begin{figure}[t]
\centering
\includegraphics[width=0.27\textwidth]{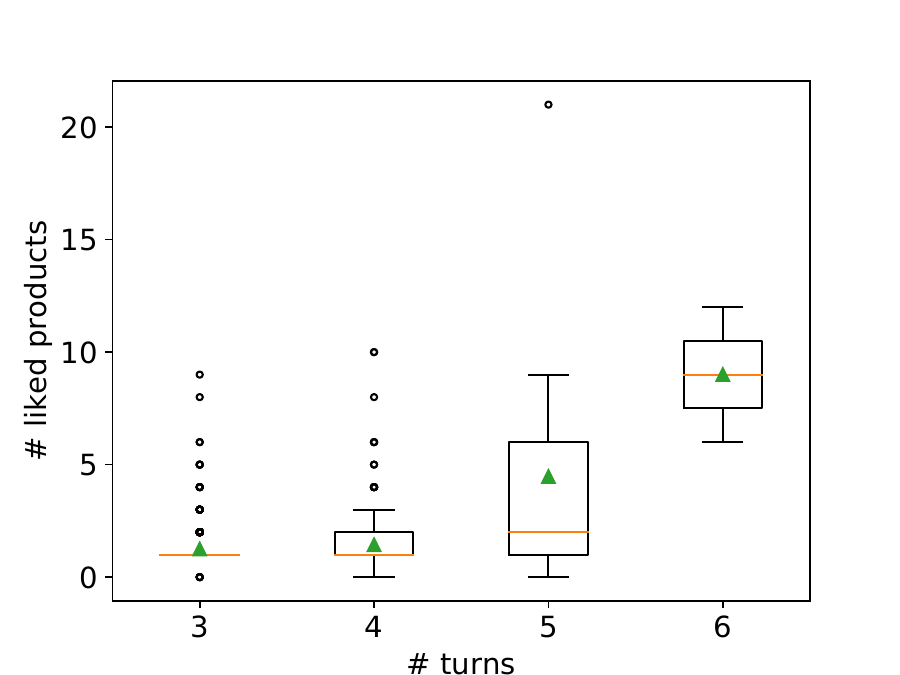}
\caption{ \# turns vs. \# liked products.}
\label{fig:box}
\end{figure}

Moreover, we explore how the number of liked products found by the system is affected by the number of conversation turns, as shown in Figure~\ref{fig:box}. We observe that a higher number of turns leads to a significantly higher number of liked products found. The one-way ANOVA test shows significant differences among different numbers of turns ($p < 0.05$). This might be because the number of recommended products increases as the conversation goes on. Also, the system captures more information and is closer to the true user preferences, and thus more likely to recommend the products the user likes, as the conversation goes on. This is in line with that the performance of \ac{CPS} models increases with the increase of the number of conversations \cite{zou2022learning}. 

\section{Task Design}
\label{sec:TaskDesign}
\noindent \textbf{User Intent Detection (T1)}
\label{sec:intent}
We define a new user intent taxonomy based on previous conversational systems~\cite{ren2021wizard,azzopardi2018conceptualizing}. The user intent taxonomy includes five intents:
(1) Reveal: reveal a new intent, i.e., issue a new query; (2) Revise: revise an intent, i.e., reformulate the query; (3) Interpret: interpret or refine an intent by answering a clarifying question or responding to product recommendations from the system; (4) Inquire: inquire about the response from the system, e.g., ask the system to rephrase its response if it is unclear; (5) Chitchat: do not fit other labels, e.g., greetings or other utterances that are not related to the user's product need. 
This user intent detection aims to detect the user intent $i_U$ for the current user utterance, given conversation history $C$ and the current user utterance $U$, which is defined as learning a mapping: ${C, U} \rightarrow i_U$.

\noindent \textbf{Keyword Extraction (T2)}
\label{sec:slotvalue}
Conversations usually contain noisy data and more content than necessary. A user utterance may also not be semantically complete due to ellipsis and anaphora~\cite{dalton2020cast}. It is challenging to feed the user's utterances and conversations to the product search model directly. Therefore, it is important to extract keywords from the user utterance for a language understanding task. 
Given the conversation history $C$ and the current user utterance $U$, this subtask aims to select a sequence of keywords $T_U$, that can best describe the user utterance. Formally, it is defined as learning a mapping: ${C, U} \rightarrow T_U$.

\noindent \textbf{System Action Prediction (T3)}
\label{sec:action}
The action stage helps the system to take appropriate actions at the right time, e.g., when to ask a clarifying question to query new information and when to recommend products. 
Considering previous work on conversational search and conversational recommender~\cite{ren2021wizard,lei2020estimation, zou2024knowledge} while making a connection to product search, we define the system actions: (1) Clarify: ask questions to clarify the user's product need for preference elicitation; (2) Recommend: give a suggestion, which can be a certain product or a list of products; (3) Inquire: inquire about the response from the user, e.g., ask the user to rephrase its response if it is unclear; (4) No-answer: cannot find related products to meet user's need; (5) Chitchat: do not fit other labels, e.g., greetings or other utterances unrelated to the user's product need. 
This subtask aims to select an optimal system action $a$, that can help the users locate their target items efficiently, given the context (e.g., conversation history $C$, the current user utterance $U$, and returned candidate items $D$). The subtask is formulated as learning a mapping: ${C, U, D} \rightarrow a$.

\noindent \textbf{Question Selection (T4)}
\label{sec:Qselect}
Asking clarifying questions is common in conversational systems to enhance the system's ability to understand the users’ underlying information needs. 
In \ac{CPS}, filtering conditions or item-related attributes can be used for generating clarifying questions \cite{zou2022learning}. 
Therefore, we cast this as human-like natural language utterances, i.e., ask a clarifying question on anchors (e.g., attributes or other slots) for preference elicitation. 
This module aims to learn to rank clarifying questions and select the top clarifying question $Q_s$ (i.e., the optimal clarifying attribute) to ask users to capture user preference, given the context (e.g., conversation history $C$, the current user utterance $U$, and candidate questions $Q$). The subtask is formulated as learning a mapping: ${C, U, Q} \rightarrow Q_s$.

\noindent \textbf{Item Ranking (T5)}
\label{sec:itemselect}
In \ac{CPS}, the system should return high-quality products to the user for selection during the conversation. If the user accepts the products, then the conversation can successfully end. 
This module aims to rank candidate items and select the top items $D_s$ for the user, based on the context (e.g., conversation history $C$, the current user utterance $U$, and returned candidate items $D$), which is defined as learning a mapping: ${C, U, D} \rightarrow D_s$.

\noindent \textbf{Response Generation (T6)}
\label{sec:response}
Like many NLP tasks, we need to generate a natural language response to the user. For example, if the system action is ``Clarify,'' the response generation module needs to generate a clarifying question. If the system action is ``Recommend,'' the response generation module needs to generate a response containing selected products. 
This module aims to translate system actions into natural language responses, given the context (e.g., conversation history $C$, the current user utterance $U$, candidate questions $Q$, and candidate items $D$) and the output of T1-T5. Formally, it is defined as a learning problem: ${C, U, Q, D, (i_U, a, Q_s, D_s)} \rightarrow Y$.

\section{Conversational Product Search}
In this section, we propose a model to replace the conversational agent for product search based on our collected data and then evaluate the model on our collected dataset.

\subsection{Base Model}
\label{sec:basemodel}
 
We adopt the Transformer framework \cite{vaswani2017attention} as our base model, consisting of the embedding and self-attention layers. 

Given a sequence $\mathcal{S}$ = $[s_1, \dots, s_k, \dots, s_K]$, the embedding of each element of the input sequence is the concentration of positional embedding $s$ and sequence embedding $p$.
\begin{equation}
\mathbf{h}_k=\mathbf{s}_k+\mathbf{p}_k.
\end{equation}

All elemental embeddings $\mathbf{h}_k$ form a matrix $\mathbf{H}$, which will then be passed through a self-attention layer. Suppose we are at the $n$-th transformer layer, we will update $\mathbf{H}$ in the following manner:
\begin{equation}
\mathbf{H}^{n+1}=\text{MultiHead}(\text{PFFN}(\mathbf{H}^{n}))~,
\label{equ:hn1}
\end{equation}
where MultiHead is a multi-head self-attention sub-layer, and PFFN is a Position-wise Feed-Forward Network constructed by the Feed-Forward Network (FFN) with ReLU activation~\cite{agarap2018deep}. Afterward, we deploy a residual connection around each of the two sub-layers (the sub-layer is MultiHead or PFFN in Equation~\ref{equ:hn1}), followed by a dropout and layer normalization. 

Given a task sequence $S$, we utilize the final hidden representations at the N-layer of Transformer $\mathbf{h}^{N}$ to generate the output of each subtask. Specifically, we apply a softmax function through a linear layer with ReLU activation in between to produce an output of probability:
\begin{equation}
\label{eq:prob}
P=\text{Softmax}(\text{ReLU}(\mathbf{W} \times \mathbf{h}^{N}+\mathbf{b}))~,
\end{equation}
where $\mathbf{W}$ is a learnable transformation matrix, and $\mathbf{b}$ denotes the trainable bias matrix.

\subsection{Task-Specific Modules}
\subsubsection{User Intent Detection (T1)}
For the task of user intent detection, the sequence is defined as $S_{T1}=[[T1], C, U]$, where $C$ is the conversation history, $U$ is the current utterance, and [T1] is an inserted classification token at the beginning of the input sequence. The sequence is passed to the Transformer base model to get the hidden representation of [T1]. Finally, the model computes the probability of user intent according to Equation~\ref{eq:prob}. 

\subsubsection{Keyword Extraction (T2)}
Similar to T1, the sequence of keyword extraction is defined as $S_{T2}=[[T2], C, U]$. After inputting the sequence into the Transformer base model, the hidden representation of [T2] is obtained. A linear classifier with a sigmoid function is utilized to predict whether each token $T$ in $U$ belongs to a keyword or not, i.e., replace softmax with sigmoid in Equation~\ref{eq:prob}. 

\subsubsection{System Action Prediction (T3)}
For system action prediction, the sequence is defined as $S_{T3}=[[T3], C, U, D]$, where $D$ denotes the candidate products. When the system takes action, the candidate products should be considered since the quality of the candidate products affects the action. In this paper, we use product titles and product descriptions as the textual information of $D$. The candidate products are products returned by the search results queried on the keywords in T2 (keyword extraction). The hidden representation of [T3] is obtained from the transformer base model and then Equation~\ref{eq:prob} is applied to predict the system actions.

\begin{table*}[tb]
\caption{Overall performance of six subtasks (T1-T6) on the collected dataset PSCon. Precision and Recall are represented by P and R in the table, respectively. Best performances are in bold.  `*' indicates significant improvements upon the best baseline in the t-test with $p$-value $< 0.05$.}
\label{tab:base}
\centering
 \small
\begin{tabular}{l|ccc|cc|ccc|cc|cc|cc}
  \toprule
  \multirow{2}{*}{\textbf{Model}} & \multicolumn{3}{c|}{\textbf{T1}} & \multicolumn{2}{c|}{\textbf{T2}} & \multicolumn{3}{c|}{\textbf{T3}} & \multicolumn{2}{c|}{\textbf{T4}} & \multicolumn{2}{c|}{\textbf{T5}} & \multicolumn{2}{c}{\textbf{T6}}\\ 
  \cline{2-15}
  & P & R & F1 & BLEU & ROUGE & P & R & F1 & nDCG & MAP & nDCG & MAP & BLEU & ROUGE \\
  \midrule
  Naive Bayes&0.53 &0.65&0.58&0.27&0.16&0.54&0.73&0.62&--&--&--&--&--&--\\ 
  Logistic Regression & 0.51 &0.62&0.54&0.25&0.16&0.57&0.75&0.65&--&--&--&--&--&--\\ 
  Random Forest& 0.57&0.63&0.59&0.25&0.15&0.62&\textbf{0.76}&0.65&--&--&--&--&--&--\\ 
  SVM&0.58&0.66&0.58&0.30&0.18&0.57&0.75&0.65&--&--&--&--&--&--\\ 
  MLP&--&--&--&--&--&--&--&--&0.54&0.38&0.29&0.11&--&--\\ 
  Llama 2 &0.52&0.28&0.21&0.47&0.49&0.66&0.28&0.26&0.27&0.08&0.25&0.08&0.18&0.21 \\
  GPT-3.5 &0.56&0.52&0.42&0.75&0.75&0.66&0.33&0.33&0.27&0.15&0.18&0.19&0.20&0.22 \\
  Llama 2 w/ finetune &0.89&0.88&0.88&\textbf{0.77}&\textbf{0.79}&0.70&0.64&0.64&0.50&0.38&0.28&0.10&\textbf{0.22}&\textbf{0.23} \\
  \midrule
  PSCon &\textbf{0.92*}&\textbf{0.91*}&\textbf{0.92*}&0.73&0.74&\textbf{0.91*}&0.64&\textbf{0.66*}&\textbf{0.70*}&\textbf{0.60*}&\textbf{0.39*}&\textbf{0.21*}&\textbf{0.22}&\textbf{0.23} \\
  \bottomrule
\end{tabular}
\end{table*}

\subsubsection{Question Selection (T4)}
For question selection, the sequence is defined as $S_{T4}=[[T4], C, U, Q]$, where $Q$ is the question. Also, we obtain the hidden representation of [T4] based on the transformer base model. 
After that, we apply a softmax function through a feedforward network with ReLU activation in between (Equation~\ref{eq:prob}) to calculate the distribution probability over candidate questions. 
In this paper, we model it as a ranking problem based on the output distribution over candidate questions, one can also model it as a binary classification problem to predict whether each candidate question is selected or not.\footnote{We experiment on it and observe a better performance when modeling it as a ranking problem than a binary classification problem on question selection (T4). The same observation holds for item ranking (T5).} 

\subsubsection{Item Ranking (T5)}
Similar to system action prediction and question selection, the sequence of item ranking is defined as $S_{T5}=[[T5], C, U, D]$. We first obtain the hidden representation of [T5] from the transformer base model. Then, similar to (T4), we apply a softmax function through a feedforward network with ReLU activation in between (Equation~\ref{eq:prob}) to calculate the distribution probability over candidate items. Again, we model it as a ranking problem. 

\subsubsection{Response Generation (T6)}
We use a standard Transformer decoder to conduct response generation. The transformer decoder outputs a representation at each decoding time step. For response generation, we put the predicted system action (T3) at the beginning to indicate the type of response. We use the representations of (T1), (T4), and (T5) as the representations of input sequences and fuse them with max pooling to get the final hidden representation. Then, a linear classifier with softmax is applied on top to predict the probability of the token at each time step. In addition, we deploy the copy mechanism~\cite{vinyals2015pointer}, which is commonly used and helpful for generating informative utterances~\cite{gu2016incorporating}, to copy tokens in the conversation history $C$, and product-related documents $D$.

\subsubsection{Training}
We use binary cross-entropy loss for keyword extraction (T2) while using cross-entropy loss for user intent detection (T1), system action prediction (T3), question selection (T4), item ranking (T5), and response generation (T6). 
To train a complex or deep model, a large number of data is needed. To this end, we first use other conversational datasets to pre-train our model and then fine-tune the model on our PSCon dataset. In our implementation, we use two conversational datasets, DuConv~\cite{wu2019proactive} and KdConv~\cite{zhou2020kdconv}, which are for knowledge-grounded conversations, for pre-training.

\section{Experiments}
\subsection{Experimental Setup}
\label{sec:expsetup}
\subsubsection{Evaluation Metrics}
The evaluation metrics used differ for different subtasks. For keyword extraction (T2) and response generation (T6), we use BLEU-1~\cite{papineni2002bleu} and ROUGE-L~\cite{lin2004rouge} as they are supposed to evaluate the quality of texts following previous work~\cite{zhao2023making}. For user intent detection (T1), and system action prediction (T3), we use Precision, Recall, and F1 as evaluation metrics as they are supposed to evaluate the quality of classification. For question selection (T4), and item ranking (T5), we use MAP and nDCG as evaluation metrics to evaluate the ranking quality of questions and items. 

\subsubsection{Implementation Details}
We run the experiments on the NVIDIA RTX A6000*8 GPU server. We use Adam optimizer~\citep{kingma2014adam} with a learning rate of 0.001 to train our model. The word embedding size and hidden size are set to 512. For Transformer, the number of layers for the encoder and decoder is set to 4 and 2, respectively. We set the Transformer head number = 8, and the global norm clip of gradients = 1 for stable training. 
We use 5-fold cross-validation to evaluate our model. 
In each fold, the dataset is split into training, validation, and testing sets by the ratio of 6:2:2. In the following experiments, we omit the results for the English dataset and only report results for the Chinese dataset due to similar trends and limited space.

\subsection{Experimental Results}
Here, we summarize the preliminary results that we achieved on the Chinese dataset. Our goal is to provide an overview of the difficulty and potential of PSCon for targeting different tasks that we introduce in this work. 
\subsubsection{Performance on Different Tasks}
We report the results of our methods on the collected Chinese dataset on all the six defined subtasks, as shown in Table~\ref{tab:base}. As there is a lack of existing conversational product search baselines that can be applied to all six subtasks, we include a set of typical and well-known classification baselines, including naive Bayes classifier, logistic regression classifier, random forest classifier, and Support Vector Machine (SVM) classifier, on the subtasks modeled by classification problems (i.e., T1-T3), while including a typical ranking baseline, Multilayer Perceptron (MLP), on the subtasks modeled by ranking problems (i.e., T4). Also, large language models (LLMs) can be employed as a solution which is a supplement to our proposed model. Therefore, we include the advanced pre-trained LLMs Llama 2 (Llama-2-7B) and GPT-3.5, by following the same setting in \citet{he2023large}, on our PSCon dataset.\footnote{
The prompts used for pre-trained LLMs: ``suppose your task is \textit{\{subtask\}}. I will
give you some contextual information. Using the contextual information as input data, predict the \textit{\{subtask output\}}. The contextual information is provided as follows: \textit{\{subtask input\}}.'' In the prompt, the \textit{\{subtask\}} is user intent detection, keyword extraction, system action prediction, question selection, item ranking, and response generation for T1, T2, T3, T4, T5, and T6, respectively. The \textit{\{subtask output\}} is user intent, keywords, system action, ranked list of questions, ranked lists of items, and response for T1, T2, T3, T4, T5, and T6, respectively. The \textit{\{subtask input\}} is shown in Section \ref{sec:TaskDesign} for each subtask.} 
The results aim to provide practitioners with a better perception of what are the advantages and challenges of using the dataset.

From the results, we observe that random forest and SVM perform the best among the four typical classification baselines for classification subtasks (T1-T3). Compared with typical classification baselines, GPT-3.5 and Llama 2 achieve relatively better performance in T2 and T6 while worse performance in T4 and T5. This might be because pre-trained LLMs like GPT-3.5 are good at natural language processing tasks (e.g., T2 and T6) but may not be good at question selection and item selection over a large number of candidate questions and items without training on the downstream datasets. With fine-tuning on the PSCon dataset, the LLM Llama 2 greatly improves its performance on the six subtasks, by achieving higher performance than GPT-3.5 and Llama 2 without fine-tuning. 

\begin{figure}[tb]
\centering
\includegraphics[width=0.45\textwidth]{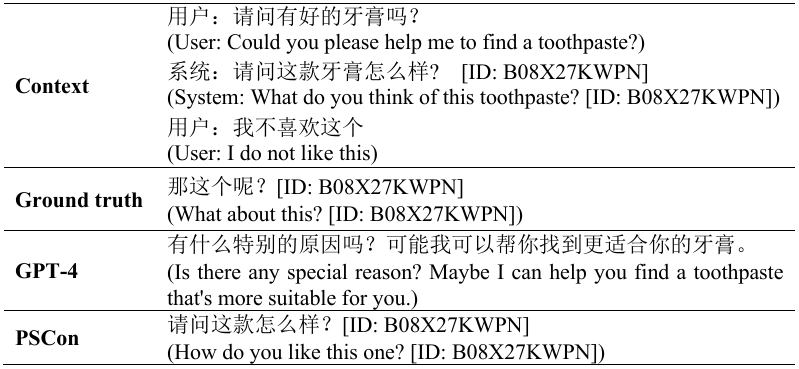}
\caption{A case study for response generation.}
\label{fig:responseexample}
\end{figure}

Moreover, we observe that the PSCon dataset is valid for examining all subtasks and the proposed model is effective for the six subtasks. Our model, although a much smaller parameter size than LLMs like GPT-3.5 and Llama 2, achieving the best performance on most subtasks, constitutes a straightforward baseline for future researchers to use. The model is designed as a unified model to model the six subtasks in a unified framework. But note that the model is not the focus of the work. 
Overall, the subtasks of user intent detection (T1), action prediction (T3), question selection (T4), item ranking (T5), and response generation (T6) achieve relatively high performance, indicating the effectiveness of our PSCon dataset on these subtasks. 
Using the conversation history for user intent detection, action prediction, question selection, item ranking, and response generation is beneficial. 
We observe higher effectiveness in terms of Precision for system action prediction, suggesting the need for further research in this area to improve recall. 
We also observe that the item ranking is challenging and far from perfect. This might be because of the large number of candidate items, which is a known problem in information filtering and recommender systems, and calls for the need to employ more complex models that learn from user-item interactions, such as collaborative filtering \cite{he2017neural}. 
Finally, we think there is a lot of room for improvement in response generation, given the low performance of GPT-3.5 and our model in terms of BLEU and ROUGE, indicating that \ac{CPS} is a promising direction but there is still a long way to go. To further validate the response generation, we add a case study shown in Figure \ref{fig:responseexample}. 
From the response generation example, we observe that, while GPT-3.5 generates general responses omitting product recommendations, our model generates reasonable responses with specific product recommendations, which are similar to the ground truth. However, the responses generated by our model do not contain the same phrase as the ground truth although their meaning is the same, leading to the relatively low BLEU and ROUGE. 

\subsubsection{Performance on User Intent Detection}
In Table~\ref{tab:intent}, we report our model's detailed performance of the intent detection task where we break down the performance in terms of each intent, namely, `Reveal', `Revise', `Interpret', and `Chitchat'.\footnote{We omit the results of the intent of `Inquire' in Table~\ref{tab:intent} and Table~\ref{tab:sysact} since there are only very few data points for the `Inquire' intent.} 
In particular, we observe that our model achieves relatively high performance for every intent, indicating the effectiveness and robustness of our model in detecting user intents. We see that our model is best at predicting `Chitchat', followed by `Reveal', exhibiting the highest Precision, Recall, and F1 for `ChitChat'. 

\subsubsection{Performance on System Action Prediction}
The detailed performance of our model in terms of each system action, namely, `Chitchat', `Clarify', `Recommend', and `No-answer', is shown in Table~\ref{tab:sysact}. 
We observe that the performance changes for different system actions. Specifically, the model achieves the highest F1 when the system actions are `Recommend' and `Clarify'. This might be because `Recommend' and `Clarify' actions are less ambiguous and can be easily classified. 

\section{Conclusion and Future Work}
\label{sec:cons}
In this paper, we introduced a pipeline of \ac{CPS}, including six sub-tasks: user intent detection, keyword
extraction, system action prediction, question selection, item ranking, and response generation. 
We collected and released a new \ac{CPS} dataset, called PSCon, based on a human-human data collection protocol, to support comprehensive and in-depth research on the six subtasks and all aspects of \ac{CPS}. The collected data is available for dual markets and two languages, and can support the development of knowledge graph-based \ac{CPS} models. We further proposed a benchmark model to model the six subtasks of \ac{CPS}. Extensive experiments on the PSCon dataset show that the model is effective. 

To collect the data, two workers need to be paired as participants to generate one conversation. Each conversation usually involves multiple searches and scanning for products, which costs a lot of effort for the participants. This makes the data collection challenging and expensive, leading to a limited volume of data like existing work
\cite{trippas2020towards,thomas2017misc,dalton2020cast,bernard2023mg}. 
We understand that, although the data contains more than one thousand conversations, the data is still limited to training a large model in an end-to-end manner. To this end, we utilize other conversational datasets to pre-train our model to relax this limitation and the experimental results of our proposed model demonstrate its effectiveness. Nevertheless, we plan to expand the dataset with more training data from a large number of users in the future. As we present a system and data collection protocol in the paper, one can easily collect more conversations by using our system and protocol. Moreover, more than one thousand conversations enable researchers to use the dataset with LLMs~\cite{mao2023gpteval} and unlock its capabilities, e.g.,  
one can employ LLM's few-shot learning by carefully curating domain-specific examples and prompts from the dataset; one can also use LLM's knowledge injection by integrating our conversations as demonstrations or leveraging our knowledge graphs through retrieval-augmented generation (RAG) to enhance its reasoning and contextual understanding. 

\begin{table}[tb]
\caption{The performance of our model for different intents for user intent detection. }
\label{tab:intent}
\centering
\small
\begin{tabular}{lccccc} 
  \toprule
 & Reveal & Revise & Interpret & Chitchat\\
  \midrule
  Precision & 0.96  & 0.83  & 0.91  & 0.98 \\
  Recall & 0.95  & 0.83  & 0.92  & 0.96 \\ 
  F1 & 0.96  & 0.83  & 0.91  & 0.97 \\
  \bottomrule
\end{tabular}
\end{table}

\begin{table}[tb]
\caption{The performance of our model for different actions for system action prediction. }
\label{tab:sysact}
\centering
\small
\begin{tabular}{lccccc} 
  \toprule
 & Chitchat & Clarify & Recommend & No-answer\\
  \midrule
  Precision & 0.73  & 0.96  & 0.96  & 1.00 \\
  Recall & 0.91  & 0.85  & 0.98  & 0.46 \\ 
  F1 & 0.81  & 0.90  & 0.97  & 0.63 \\
  \bottomrule
\end{tabular}
\end{table}

In this paper, we experiment with our PSCon dataset to offer an overview of the difficulty and potential of our PSCon for targeting different tasks. Nevertheless, we do not experiment with cross-lingual or cross-market models \cite{zhu2022cross, bonab2021cross} on our datasets. As we collect two datasets from two markets with two languages, there are several items from the international e-commerce stores that are the same across the two datasets. One can use the conversation in one market with a certain language to retrieve the target products in the other market with a different language. The information and knowledge of items (e.g., the conversation, product metadata, and product reviews) can be transferred from one dataset to another dataset, which can be used for training domain-adaptation product search models. 
Also, the item interactions from conversations in one dataset can be learned and used as collaborative signals (e.g., users like Roccat Kain 120 AIMO mouse may also like Roccat Kone XP Air mouse, as stated in our PSCon dataset) in the other dataset for improving cross-domain (i.e., cross-lingual or cross-market) product recommendations. We plan to expand the dataset with more languages/markets and train cross-domain product search models (e.g., \cite{bonab2021cross} and \cite{zhu2022cross}) on our datasets in our future work. 


Although we have designed six subtasks for \ac{CPS} and proposed a Transformer model to model these six subtasks, the proposed model is straightforward and simple. It constitutes a straightforward baseline for future researchers to use. Nevertheless, it is worth mentioning again that \ac{CPS} is a promising direction and there is still a long way to go to improve the performance of each of the six subtasks. 

\begin{acks}
This research was supported by the National Natural Science Foundation of China (62402093) and, the Sichuan Science and Technology Program (2025ZNSFSC0479). This work was also supported in part by the National Natural Science Foundation of China under grants U20B2063 and 62220106008, and the Sichuan Science and Technology Program under Grant 2024NSFTD0034.
\end{acks}

\clearpage

\bibliographystyle{ACM-Reference-Format}
\bibliography{bibfile} 

\end{document}